\RequirePackage{fix-cm}
\documentclass[twocolumn]{svjour3}          % twocolumn
\smartqed  % flush right qed marks, e.g. at end of proof
\usepackage{graphicx}
\usepackage{multirow}
\usepackage{array}
\usepackage{booktabs}
\usepackage{longtable}
\begin{document}
%============ Paper Title =================
\title{3D Ultrasound image segmentation: A Survey}
%============ Authors name ===============
\author{M. Hamed Mozaffari \and WonSook Lee
}
\institute{M. Hamed Mozaffari \and WonSook Lee \at
              School of Electrical Engineering and Computer Science, University of Ottawa, 800 King Edward Avenue, Ottawa, ON K1N 6N5, Canada \\
              Tel.: +123-45-678910\\
              Fax: +123-45-678910\\
              \email{mmoza102,wslee@uottawa.ca} 
}
\date{Received: date / Accepted: date}
\maketitle
%============= Abstract ===================
\begin{abstract}
Three-dimensional Ultrasound image segmentation methods are surveyed in this paper. The focus of this report is to investigate applications of these techniques and a review of the original ideas and concepts. Although many two-dimensional image segmentation in the literature have been considered as a three-dimensional approach by mistake but we review them as a three-dimensional technique. We select the studies that have addressed the problem of medical three-dimensional Ultrasound image segmentation utilizing their proposed techniques. The evaluation methods and comparison between them are presented and tabulated in terms of evaluation techniques, interactivity, and robustness.

\keywords{3D Ultrasound imaging \and 3D Ultrasound image segmentation \and Medical image processing}
\end{abstract}

%=========== Introduction ===================
\section{Introduction}
\label{intro}
For many diseases such as cancers, physicians have to see one region of the body with a specific issue to diagnose illness, estimate the current situation then can properly decide which treatment is the best.  The information acquired from the body is also critical for the surgeons to understand the exact position of the defect and direct their instruments to that position accurately. So, image processing methods can assist practitioners to find the region of interest easier. 
\newline Ultrasound (US) is one of the image modalities which is designed for monitoring the soft tissues. Interpretation of US images is always a challenging task due to the noise, shadows and artifacts in the images. In medicine, it is always favorable to delineate one organ of the body especially whose have a problem in order to better investigate its condition. 
\newline Image segmentation is absolutely essential preparatory process in almost all image processing approaches. The goal of this methods is to delineate the region of interest (ROI)(foreground) from the image to use in another post processing procedures accurately, automatically and in real-time. Many techniques are presented for image segmentation and working in this field still is a hot topic. As a common classification, image segmentation techniques can be considered as Thresholding, Region Growing, Deformable Surfaces, Level Set methods, Graph-based techniques and sate of the art approaches [1]. 
\newline Although, we can extract lots of information from 2D images and its segmentation is significantly useful for many applications but for medicine, understanding the 3D images is easier than 2D images. From 3D images we can understand more details about the problem. For capturing and interpretation of 3D image, huge amount of storage memory, and processors are needed. 3D US image capturing using 3D probes are expensive and also impossible for big organs. Recent developments of computer reconstruction algorithms able the researchers to reconstruct 3D US images from 2D images and work with 3D and also 4D images. Many studies have been done on 3D image reconstruction and 2D US image segmentation but its out of scope of this review. 
\newline In this paper, we focus on just the studies of Ultrasound image segmentation in 3D domain and we have not included 2D methods. Indeed, we used the way of categorization in [2] for 3D US image segmentation techniques in terms of clinical applications. To our knowledge, this is the first review in this shape and we tried to gather all the papers in each cluster from the literature. We also try to briefly explain principle concepts of image segmentation in each clinical class. 
\newline The rest of this paper is organized as follow: In section 2, methods are divided into medical applications including, Prostate cancer, Breast Cancer and methods are used for Needle, Kidney, Embryo and Fetus, Cardiovascular and Carotid arteries, and other techniques. Section 3 conclude the paper.

%======================================
%=========== Clinical Application Section ==========
\section{Clinical Applications as a Taxonomy}
\label{sec:1}
In this section, we have classified 3D Ultrasound image segmentation methods with respect to their application in Medical and Clinical purposes. Taxonomy is designed by the number of paper topic repetitions in each area. For papers which work on other specific parts of the human body, we gathered them in miscellaneous category.  
%============ Prostate Cancer ================
\subsection{Prostate Cancer}
\label{sec:2.1}
All over the world, the second most common cancer diagnosed and the sixth most common cause of cancer death among men is prostate cancer \cite{ProstateStatistic}, \cite{paper1}. Early diagnosis of this cancer is vital and patients survival as a result. Having a 3D simulated structure of prostate and cancerous tissues would be helpful for physicians to better selection of therapeutic modality and better localization of problematic tissues and seed implantation which are used by surgeons in real-time planning. Prostate brachytheraphy quality assessment cause costs reduction and healing result increment \cite{paper130}.
Therefore, accurate and reproducible human prostate segmentation from Ultrasound images is a crucial step of many diagnosis and treatment procedures for prostate diseases \cite{paper122}. 
\newline To segment the prostate boundary either automatically or semi-automatically from 3D US images a number of algorithms have been developed. This object still is a challenging procedure because of the low contrast and worse quality of prostate US images which user knowledge is considered as initialization step for overcome to this difficulties in many techniques. 
%deformable 
\newline Ghanei et al. \cite{paper122} applied a 3D discrete deformable surface for accurate outline of prostate by using bilinear interpolation after acquiring the 2D ultrasound images. The operator need to draw an inaccurate closed polygons with four or five points for some of the slides (about 40\% - 70\%). The proposed method, deforms the user initial model by movement of its vertices with defining two forces. The internal forces try to maintain the smoothness of the model by minimizing the surface curvature using least squares error estimation of the Dupin indicatrix. The external Forces which are extracted from the image features tries to pull the model toward the prostate boundaries. 
To initialize deformable model method in ref \cite{paper49} six control points are selected by operator to estimate a posterior 3D prostate shape as a triangles mesh. Then, the mesh deforms to localize the prostate boundary by applying the forces that propels points on the mesh toward edges and using a simulated surface tension for keeping the mesh smoothness. Because of image-based forces affect points in short distances, editing process also is applied in the case that some mesh points are far from the prostate boundaries due to incorrect user initialization. 
Due to different user initializations for the same subject in different runs of the algorithm in \cite{paper1} authors also studied and added variability and accuracy evaluation of their previous research final results \cite{paper49} by calculating standard deviation of the mesh distribution and average of meshes.  
\newline Wang et al. \cite{paper120} reported using of two methods for 3D US segmentation which in both 3D prostate data was sliced into uniform parallel adjacent images and rotational form around a common axis (approximately prostate center). The second method is suitable for round objects, such as the prostate. 
\newline Discrete Dynamic Contour (DDC) is an extended version of deformable model used in \cite{paper124} and \cite{paper120} to refine the initial boundaries which instead of minimizing the whole contour energy attempts to find the optimum energy of contour vertices. Authors in \cite{paper124} used the Cardinal-spline as an interpolation function to find initial boundary from vertices and in similar modality \cite{paper123} they adding a continuity constraint by using an autoregressive (AR) model for better approximation.
\newline A fast version of segmentation method using AR model and a continuity constraint was implemented by similar research group in \cite{paper129}. Ladak et al. \cite{paper21} proposed a 3D deformable model as a Volume-based method which is represented by a closed mesh of triangles connected at their vertices, similar to the straight lines in slice-based methods. Again here weighted internal, external and damping forces apply to user defined initialized mesh to deform it toward the border of the Prostate cancer.  
%level-set
\newline Limitation of deformable methods such as boundary leaking due to low quality images, weak performance of the method in complex geometry and complicated implementation addressed by using level-set modalities \cite{paper125}. Authors in \cite{paper127} defined a new energy formulation for prostate shape visualization which is obtained from a combination of shape and intensity prior knowledge in a level set framework with a Bayesian interface. They presented an automatic 3D US segmentation technique by minimization of this energy function without special initialization. Qiu et al. \cite{paper121} utilized rotational slicing technique of 3D US data to 2D images for delineation of prostate regions and employed level set method with shape constraint to reconstruct the prostate shape. Level set function was used with local-region-based energies to refine the weak regions and edges of prostate boundaries. 
The similar research group \cite{paper131} introduced a new convex optimization-based approach to obtain the prostate surface from a given 3D US image. They sliced 3D data to rotational images and manually set initial prostate boundary and central points on the coronal and transverse view. Then using this points optimization algorithm find the prostate contour and propagate it to other slices for reconstruction of 3D structure. \newline Hodge et al. \cite{paper53} used 2D Active Shape Models (ASM) with rotation-based slicing of 3D US data and in \cite{paper6} they used an optimization technique for similar ASM method to find the optimum 2D point distribution model (PDM). Fan et al. \cite{paper125} demonstrated the performance of a fast level-set technique to solve the boundary leaking problem.  
%statistical shape model
\newline Shen et al. \cite{paper126-ref14} proposed a statistical shape model using Gabor filter bank to characterize the prostate boundaries. They first found the statistical shape of the prostate from manually outlined training samples and also from testing sample which Gabor features are obtained in this step, then using Hierarchical deformable technique and training information, prostate shape is segmented from the transrectal ultrasound (TRUS) data. The same research group \cite{paper126} gathered texture priors information as well as statistical shape data in the training stage and used that for training a Gabor Support Vector Machine (GSVM). Prostate tissues voxels from non-prostate once separated by Hierarchical deform model using GSVM in the next stage. 
\newline Some authors used a modified and mixed statistical shape methods to find the best results such as Heimann et al. \cite{paper29} introduced a method of Statistical shape models with some specific appearance models (Gaussian Gradient, Non-linear Gradient, local histogram) to match prostate shape with image data. A novel combination of Statistical model-based automatic segmentation as an constraint optimization problem is designed by Shao \cite{thesis116} which try to find the best shape and pose parameters between training samples to discriminate prostate tissue from background.
\newline Recently, Prostate cancer tissues are diagnosed by TRUS-guided biopsy as an standard method and having an accurate biopsy technique still is a challenge and open area for researchers.  
Yang et al \cite{paper45}, \cite{paper35} proposed a 3D segmentation of prostate TRUS using multi-atlas and longitudinal image registration. As initialization and learning step they registered and manually segmented all the 3D US data, then matched intensity of newly acquired images with the database images using histogram matching algorithm. Three orthogonal Gabor filter banks as a texture feature extraction are applied to database and new images. combination of the output results are used in training stage of kernel support vector machines (KSVMs). So, new prostate images are delineated by the trained KSVMs. In \cite{paper31}, \cite{paper2} similar automatic segmentation technique performance is demonstrated by adding a statistical shape model, intensity profiles and texture information to a set of Wavelet-based support vector machines (W-SVMs) along with training samples.  Authors use wavelets in 3 cross section planes for texture extraction of the prostate regions and also use a probability model for enhancement of accuracy and robustness of the method. Prostate and non-prostate tissues are classified around their boundaries in the next step.
\newline In \cite{paper130} a new optimization method is employed to fit the best surface of the prostate to the underlying images under shape constraints. Spherical harmonics of degree eight used for modeling of the prostate shape as constraints and statistical analysis performed on the shape parameters. Mahdavi et al. \cite{paper128} presented a method using shape enhancement algorithms such as un-wrapping, un-tapering and mid-gland ellipse fitting to find the actual shape of prostate which has been changed due to the presence of the TRUS probe with the help of perior prostate shape knowledge. \newline Nouranian et al. \cite{paper37} proposed a new automatic algorithm using prostate 3D image atlases as priori knowledge. Authors after registration and segmentation of the atlases applied the Simultaneous Truth And Performance Level Estimation (STAPLE) algorithm to find the probabilistic estimate of the true segmentation in each atlas data and then used outcome to find the best shape matching for new target image. Fenster et al. \cite{paper86} described basic ideas and some well-known techniques of 3D segmentation and visualization of the prostate, needle and seeds to be used in 3D US-guided prostate brachytherapy. 
\newline Some of 3D prostate segmentation methods summary is tabulated in table \ref{table1}. Key words of the table are: A: Automatic, M: Manual, MD, MAD and MAXD are Mean Difference, Mean Absolute difference and Maximum difference between manual results and algorithm results in millimeters respectively. The percent volume difference (PVD)(in percentage or $cm^3$) and the percent absolute volume difference (PAVD) are calculated to assess global performance of the methods. AM, DAM and SD are average segmentation for each set of meshes which are determined by specific number of algorithm repetitions, difference between each individual mesh and average one and the standard deviation around the mesh boundaries respectively. For finding significant differences in results t-test was used. OE: volumetric overlap error, ASD: average surface distance, RMSD: rms value of surface distance and MSD: maximum surface distance, DSC :percentage of dice similarity (overlap) ratio (coefficient), SEN: percentage of sensitivity, FNR: False negative rate, ER: Error ratio, VE: volume error, $R_c$ is the relative amount of agreement (similarity value), F-test: F statistics to compare the variance estimations, CSR: Correct Segmentation Rate, ISR: Incorrect Segmentation Rate. Note that all comparisons are done between the gold standard (manual results) and algorithm results. 
\begin{figure}
\includegraphics[width=0.5\textwidth]{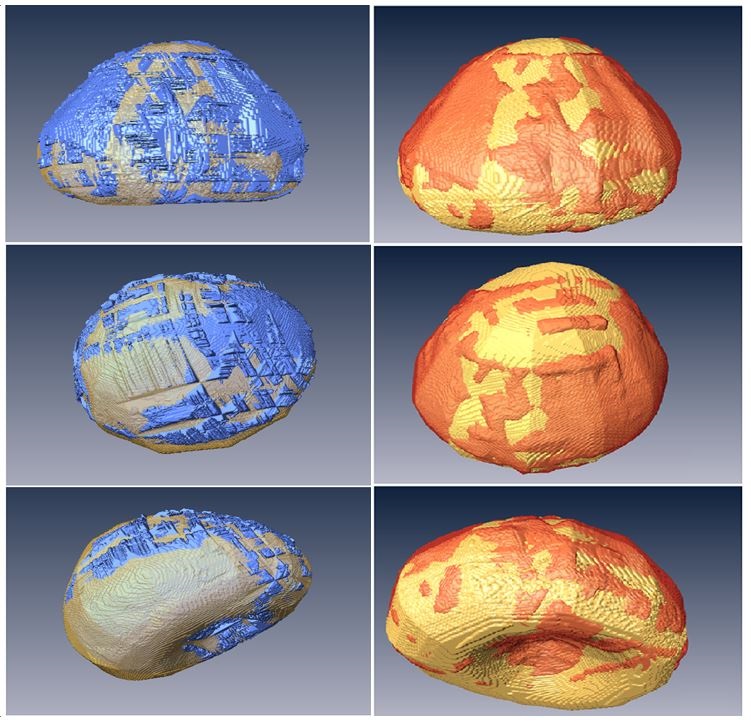}
\caption{Three views of 3D prostate segmentation visualization (shown in blue \cite{paper31} and red \cite{paper2}) and its comparison with the manual gold standard (shown in gold)}
\label{fig:1}       % Give a unique label
\end{figure}
%----------  Table 1 -----------------
\begin{table*}[]
\caption{Prostate Cancer segmentation methods}
\label{table1}  
\begin{tabular}{m{1.25cm}cm{2.5cm}cm{3cm}m{3cm}m{3.5cm}}
\hline
\noalign{\smallskip}
%-----------------
Reference                      &Year         &Modalities                     &A/M &Samples Acquisition                    
&Evaluation Methods                                                                
& Values\\ \hline\noalign{\smallskip}
%-----------------
Ghanei \cite{paper122}  &2001       &3D deformable model                                       &M    &10 TRUS 3D volumes 
&$R_c$
&0.89 \\ \noalign{\smallskip}
%-----------------
Fan \cite{paper125}      &2002       &Level-set Method                                               &M  &8 TRUS 3D volumes
&-
&-\\ \noalign{\smallskip}
%-----------------
Hu \cite{paper1}\cite{paper49}           &2003         &Deformable Model        &M     &6 mechanical B-mode transducer rotational probe data 
&MD, MAD, MAXD, PVD, AM, DAM, SD, t-test  
& -0.20$\pm$0.28, 1.19$\pm$0.14, 7.01$\pm$1.04, 7.16$\pm$3.45\% ,~ ,~ ,~ ,51.5\%\\ \noalign{\smallskip}
%-----------------
Ding \cite{paper124}     &2003       &Deformable Sliced based models + DDC + Cardinal-Spline interpolation &M &6 TRUS 3D volumes, acquired by tilt scanning mechanism
&PAVD, 
&4.53\%\\ \noalign{\smallskip}
%-----------------
Wang \cite{paper120}   &2003        &ASM + Discrete Dynamic contour (DDC) (2 methods)   &M   &6 TRUS 3D volumes in vivo using a tilt
motorized scanning mechanism
&PVD, PAVD
&- \\ \noalign{\smallskip}
%-----------------
Ding \cite{paper123}     &2004       &Constrained Deformable sliced-based models     &M    &6 TRUS 3D volumes, acquired by tilt scanning mechanism
&MD, SD
&-\\ \noalign{\smallskip}
%-----------------
Ladak \cite{paper21}    &2003         &3D Deformable Model (3 algorithms)  &M    &4 TRUS acquired 3D images in vivo 
&PVD
&5.01\%\\ \noalign{\smallskip}
%-----------------
Tutar \cite{paper130}    &2006       &Parametric Deformable Models + Optimization Algorithm   &M &30 TRUS 3D volumes
&MAD, MAXD, OE
&1.26$\pm$0.41, 4.06$\pm$1.25, 83.5$\pm$4.2\% \\ \noalign{\smallskip}
%-----------------
Yang \cite{paper127}    &2006       &Active Shape and Intensity priors Models + Energy Optimization    &A &11 TRUS 3D rotational reconstructed volumes
&CSR, ISR
&0.82$\pm$0.05, 0.19$\pm$0.08\\ \noalign{\smallskip}
%-----------------
Zhan \cite{paper126}    &2006       &Statistical Shape Model + SVM + texture and Gabor   &A &6 TRUS 3D volumes
&ASD(Average Distance), OE, PVD (Volume Error)
&1.12$\pm$0.15 (voxels), 4.16$\pm$0.54\%, 2.22$\pm$1.19\%\\ \noalign{\smallskip}
%-----------------
Hodge \cite{paper6}\cite{paper53}     &2006         &2D Active Shape Model &M     &36 volumetric images using a 3D TRUS imaging system
&MD, MAD, MAXD, PVD, PAVD
&0.12$\pm$0.45, 1.09$\pm$0.49, 7.27$\pm$2.32, 0.22$\pm$4.58\%, 3.28$\pm$3.16\%\\ \noalign{\smallskip}
%-----------------
Ding \cite{paper129}     &2007       &Constrained sliced-based autoregressive model     &M &9 TRUS 3D volumes
&t-test, F-test, MAD, SD, AM, DAM
&- \\ \noalign{\smallskip}
%-----------------
Heimann \cite{paper29}   &2011     &Statistical Shape Model (3 algorithms) &A     &35 TRUS 3D volumes 
&OE, ASD, RMSD, MSD
& - \\ \noalign{\smallskip}
%-----------------
Mahdavi \cite{paper128}  &2011    &Shape Model fitting using 3D enhancement shape modification  &M &-
& PAVD, PVD, MAD, MAXD, SEN
&- \\ \noalign{\smallskip}
%-----------------
Yang \cite{paper45}     &2011         &Statistical texture and Atlas-based Model + KSVM  &A  &5 TRUS 3D volumes
&DSC, OE, ASD, RMSD, MSD
&90.81$\pm$1.16\%, 16.44$\pm$1.93\%, 1.61$\pm$0.35, 1.72$\pm$0.47, 5.04$\pm$1.26 \\ \noalign{\smallskip}
%-----------------
Akbari \cite{paper31}   &2011         &KSVM + Statistical Shape Model  &A &5 TRUS 3D volumes
&DSC, SEN
&90.7\%, 4.9\%  \\ \noalign{\smallskip}
%-----------------
Akbari \cite{paper2}     &2012         &Statistical Shape Model + texture and intensity info. + 4 SVMs  &A &40 TRUS image volumes of 20 patients
&DSC, SEN, FNR, OE, VE
&90.3$\pm$2.3\%, 87.7$\pm$4.9\%,~,~,~ \\ \noalign{\smallskip}
%-----------------
Yang \cite{paper35}     &2012         &Longitudinal Registering + Statistical texture Model+ KSVM &A &5 TRUS 3D volumes
&DSC, ASD, RMSD, MSD
&88.1$\pm$1.44\%, 1.18$\pm$0.31, 1.43$\pm$0.31, 3.89$\pm$0.7 \\ \noalign{\smallskip}
\end{tabular}
\end{table*}
\begin{table*}[]
\begin{tabular}{m{1.25cm}cm{2.5cm}cm{3cm}m{3cm}m{3.5cm}}
%-----------------
Qiu \cite{paper121}     &2013         &Level-set Method + shape constraints                &M    &35 TRUS 3D volumes using rotational scanning system
&SEN, DSC, PVD, MAD, MAXD
&93.0$\pm$1.6\%, 93.1$\pm$2.0\%, 2.6$\pm$1.9, 1.18$\pm$0.36\, 3.44$\pm$0.8 \\ \noalign{\smallskip}
%-----------------
Nouranian \cite{paper37} &2013     &Atlas based Statistical Model + STAPLE algorithm   &A  &50 TRUS 3D volumes
&PVD, OE, DSC
&-3.61$\pm$9.4\%, 8.72$\pm$2.49\%, 91.28$\pm$2.49\% \\ \noalign{\smallskip}
%-----------------
Qui \cite{paper131}      &2015        &Sliced-base Convex Optimization Algorithm    &M   &30 patient TRUS 3D volumes with rotational scanning biopsy system
&DSC, SEN, MAD, MAXD
&93.4$\pm$2.2\%, 92.6$\pm$2.8\%, 1.12$\pm$0.4, 3.15$\pm$0.65\\ \noalign{\smallskip}
%-----------------
\noalign{\smallskip}\hline
\end{tabular}
\end{table*}

% -------- end table 1 --------------------

%============ Breast Cancer ================
\subsection{Breast Cancer}
\label{sec:2.2}

According to \cite{BreastCancerStatistics} the deadliest cancer among women is Breast Cancer. Although, mammography is widely used for finding breast cancer but Ultrasound imaging is another useful, inexpensive, non-invasive, pervasive modality and sometimes as a supplementary for mammography to find the cancer tissues positions. Determining location of breast tumor, size and shape of it from Ultrasound images is very important for physicians to make an accurate diagnosis and treatment. So precision of the segmentation algorithm is significantly affects on tumor volume finding.
\newline Chen et al. \cite{paper7} designed a computer program for breast cancer segmentation using the Discrete Dynamic Contour Model (DDCM) and was attempted to find the breast volume from initial contours curves, then 3D VIEW 2000 results was used for comparison. Edge information, after applying image processing techniques to 2D images such as blurring, thresholding, opening and closing have been used to find initialized boundary. 
\newline In \cite{paper17} authors applied region-based image processing techniques such as split-and-merge and seeded region growing using a distortion-based homogeneity to find homogeneous regions of the images as tumor segments. 
\newline To extract 3D shape model of breast cancer from 3D US data, Chang et al. \cite{paper133} used 3D Snake Models (Active Contour Models). They first applied anisotropic filter and thresholding method to find initial shape of the cancer region. They applied 3D snake procedure as delineation method in the next step. In \cite{paper134} they have done similar research but after 3D data acquisition process, initialized 3D shape model is found as binary images, calculated by thresholding, closing and opening techniques. Kuo et al. \cite{paper47} utilized radial gradient index (RGI) which is a seeded segmentation algorithm to find initial contour of the breast cancer and then proposed an active contour based delineation method. Liu et al. \cite{paper27} first used anisotropic diffusion filter to remove noises and enhance image contrast, then a mathematical morphology process was executed to find initialization borders for each 2D slides. For segmentation of each 2D ultrasound image individually, they applied the level-set method and put together the resulting contours to form a 3D representation of the tumor boundary.
\newline In \cite{paper55} a general back propagation learning multi-layered perceptron (MLP) neural network and some of image enhancement techniques such as sigmoid filters, local variance enhancement filters and stick algorithm diffusion were utilized for tumor shape finding from image background. In this method for a sequence of 2D slices, each slice is consider as a reference image for the next one by extracting five image features. Gu et al. \cite{paper5} proposed a new method using Sobel operator and watershed transform to find edge information from gradient magnitude images. They applied a morphological image enhancement to reduce noises and region classification before and after the segmentation process respectively.
Hopp et al. \cite{paper43} utilized a new method for (semi-)automatic segmentation of Ultrasound Computer Tomography (USCT) breast cancer using slice-wise Canny edge detection algorithm and 3D surface fitting for smoothing enhancement. 
\newline Statistical algorithm such as EM-MPM (Expectation Maximization with Maximization of Posterior Marginals) have been adopted for segmentation in \cite{paper113} and \cite{paper132}. Yang et al. compared two segmentation methods (EM-MPM and K-means Clustering) in \cite{paper113}. They also in \cite{paper62}, \cite{paper62-2} compared and analyzed the ability of two similar techniques for 3D US breast cancer segmentation, the Bayesian algorithm using EM-MPM which is a classifier for finding the best similar pixels probabilistic of image and K-means clustering that segment image to regions as k clusters. As a conclusion from their studies they found that EM-MPM acts better than K-means Clustering. In table \ref{table2}. a brief review of the methods is illustrated and table keys are: SR: Similarity Rate, ER: Error rate, OR: Overlap Ratio, TP: True Positive, FP: False Positive, FN: Negative Positive, PDA: Percent Density Assessment, SE: Segmentation Error, JSI: Jaccard Similarity Index, RMSE: Root Mean Square Error. Note that all comparisons are made between manual and proposed algorithms results. 

%----------  Table 2 -----------------
\begin{table*}
\caption{Breast Cancer segmentation methods}
\label{table2}  
\begin{tabular}{m{1.25cm}cm{4.5cm}m{4.5cm}m{2cm}m{2cm}}
\hline
\noalign{\smallskip}
%-----------------
Reference                      &Year         &Modalities                               &Samples Acquisition                    
&Evaluation Methods                                                                
& Values\\ \hline\noalign{\smallskip}
%-----------------
Chang \cite{paper134}  &2003         &3D Active Contour Models + thresholding, 3D morphology, closing and opening techniques  &8 3D US data
&SR (Match Rate)
&95\%\\ \noalign{\smallskip}
%-----------------
Chang \cite{paper133}  &2003         &3D Active Contour Models + anisotropic filter and thresholding techniques      &4 3D US data from Mechanical tilt transducer 
&-
&-\\ \noalign{\smallskip}
%-----------------
Kwak \cite{paper17}     &2003         &Region-based method              &2 US volume data: real and artificial
&ER
&17\% \\ \noalign{\smallskip}
%-----------------
Chen \cite{paper7}        &2003        &Deformable Model + DDCM      &8 image sequencees of tumor volumes
&SR
&- \\ \noalign{\smallskip}
%-----------------
Liu \cite{paper27}       &2007          &Level-set method + anisotropic filter       &1 freehand 3D US data
&-
&-\\ \noalign{\smallskip}
%-----------------
Huang \cite{paper55}  &2008          &MLP Neural Network + Image Enhancement methods   &94 (23 benign cases and 71 malignant cases) 3D US images
&TP, FP, FN
&23 benign: 45.4\%, 27.0\%, 54.6\%, 71 malignant: 66.1\%, 17.3\%, 33.9\% \\ \noalign{\smallskip}
%-----------------
Yang \cite{paper113} &2013          &EM-MPM + K means Clustering                        &20 3D US of synthetic phantom ultrasound tomography
&PDA
&-\\ \noalign{\smallskip}
%-----------------
Kuo \cite{paper47}      &2013          &Active Contour + radial gradient index    &98 3D breast ultrasound images
&OR
&-\\ \noalign{\smallskip}
%-----------------
Hopp \cite{paper43}  &2014           &Slice-wise edge detection + Surface fitting      &16 in-vivo 3D US datasets acquired in the 3D USCT           
&Compared with MRI images, RMSE
&-\\ \noalign{\smallskip}
%-----------------
Gu \cite{paper5}         &2016          &Edge-based method + Region Classification    &21 3D data from dual-sided automated breast ultrasound system 
&OR, PDA, SE, JSI
&85.7\%, 86\%, 8.1\%, 74.5\%\\ \noalign{\smallskip}
%-----------------
\noalign{\smallskip}\hline
\end{tabular}
\end{table*}
% -------- End table 2 --------------------

%============ Needle ================
\subsection{Needle}
In some clinical application and Therapeutic methods for cancers like prostate brachytherapy, physicians have to insert needles into the prostate tumors. The US beams in almost all cases is approximately perpendicular to the tumor especially when surgeon is dropping radioactive seeds. Trajectory of needle tip is vital to reach to the correct position in the tumor. So the needles must be inserted along the right and accurate direction with a pre-planned path and insertion stopping specification. Using 2D B-scan Ultrasound for needle path finding is very common but challenging because of the noise and limitation of 2D transducer space visualization. Also, in most cases the orientation of the prob and needle are the same and finding the optimum route from 2D image become harder and is highly dependent on the skill of the physician. Therefore, 3D reconstruction of 2D images and segmentation of the needle position before and after its injection would be more accurate and useful in applications. Specially when there is a significant error (over 5 mm) and the operator have to withdraw and reinsert the needle. In this conditions, a 3D simulation of the actual needle trajectory could be achieved by a rapid re-planing, segmentation and updating the desire needle position. Drawback is huge amount of data for processing with speckle noises and shadowing which opens an area of research for computer scientists.
\newline Ding et al. \cite{paper16}, \cite{paper79} proposed a method for needle segmentation using 2D projection planes associates with needle direction from 3D US data and an adaptive 1D search technique which crops needle trajectory from the 2D slices parallel, perpendicular to the needle. The cropped volume is rendered with Gaussian transfer functions to 3D volume. They also found \cite{paper118} that the 3D vector describing the needle direction lies along two orthogonal planes to the projection direction and the needle direction in the projected 2D image would be reduced the task of 3D needle segmentation to two 2D needle segmentations. 
\newline Needle as a remedial technique also is called brachytherapy. It is used for destroying prostate cancer cells which is a process of inserting radiation seeds in the patient's prostate. So the radiation dose amount, exact place of the seeds and place of the needle tip during the insertion is critical. Ding et al. in \cite{paper4} used similar projection method for 3D TRUS images in order to segment brachytherapy needle and seeds from prostate tissue. 
\newline Zhou et al. \cite{paper28} used the Hough Transform (HT) and the 3D Randomized Hough Transform (3DRHT) which are 2D line-detection techniques are applied to 3D segmentation of needle in 3D US data space. They utilized their method on finding needle position and orientation of invasive ablation system for Uterine adenoma and bleeding which are the two most common diseases in woman. They also used a 3D Improved Hough Transform (3DIHT) algorithm based on coarse-fine search strategy and volume cropping in \cite{paper44}. Using the Hough Transform Approach for segmentation of needle in prostate cancer biopsy was also investigated by Hartmann et al. \cite{paper114}. They first applied a thresholding filter to reduce mistakes and classification algorithm attempts to find the brightest and longest line in images as the needle direction.
Qui and Ding \cite{paper138} proposed a new 3D Quick Randomized Hough Transform (3DQRHT) by adding coarse-fine search strategy to 3DRHT in order to address real time problem of 3DHT and 3DRHT methods. 
\newline Authors in \cite{paper56} demonstrated a new modality based on parameterization of the shape of the needle (specially curved needles) using Bézier curves and the generalized Radon/Hough transform (GRT) for real-time detection of curved needles in 3D and increased the speed of calculations by using graphics processing unit (GPU). A projection of a noise filtered 3D US image onto a 2D image for segmenting curved needle was used in \cite{paper136}. Wei et al. \cite{paper137} proposed a new methods of needle position segmentation in Transperineal Prostate Brachytheraphy. Their 3D TRUS-guided system scans two times. Volume data before and after the needle insertion are captured, then Grey-level change detection technique finds the differences between the prior without needle and acquired needle contained images. 
\newline Zhao et al. \cite{paper30} presented a novel 3D modality called 3D Phase-grouping which is a 3D version of Brian Burns extraction technique \cite{paper30-ref5} with the gradient orientation and general intensity variations associated with that straight line for segmenting straight line. For curved needle segmentation, Adebar et al. \cite{paper102} presented a new approach in 3D US by applying external vibration to the needle and detect its location by using Doppler vibration position imaging. The similar research group used combination of the B-mode US images and Doppler image for vibration detection (for identifying the regions of interest in the B-mode images) utilized in \cite{paper93}. Table \ref{table3}. is a brief review of the needle segmentation methods and note that the results were compared with gold-standard data.

%----------  Table 3 -----------------
\begin{table*}
\caption{Needle segmentation methods}
\label{table3}  
\begin{tabular}{m{2cm}m{1cm}m{3cm}m{2.5cm}m{4cm}m{3cm}}
\hline
\noalign{\smallskip}
%-----------------
Reference                      &Year         &Modalities                               &Samples Acquisition                    
&Evaluation Methods                                                                
& Values\\ \hline\noalign{\smallskip}
%-----------------
Ding \cite{paper16} \cite{paper118} \cite{paper79} \cite{paper4}        &2002, 2003, 2004, 2006        &1D Search algorithm + 2D Image Projection, Volume Rendering and Cropping     &Agar and Turkey breast phantoms 3D US data
&Effect of method parameter variation on the results, position and orientation accuracy analysis
& \\ \noalign{\smallskip}
%-----------------
Wei \cite{paper137}       &2004     &Grey-level change detection    &3D TRUS-guided and robot-assisted data of the chicken tissue and agar phantoms
&Robot accuracy and speed
&- \\ \noalign{\smallskip}
%-----------------
Zhou \cite{paper28}  &2007       &3DHT and 3DRHD  &3D US captured from water phantom 
&Accuracy of orientation deviation, position deviation, speed
& 3DHT: 1.30 $^{\circ}$, 1.76 mm, 3.05s, 3DRHD: 2.62 $^{\circ}$, 2.39 mm, 0.10s \\ \noalign{\smallskip}
%-----------------
Qui \cite{paper138} &2008  &3DQRHT  &Water phantoms 3D US images
&Accuracy of orientation deviation, position deviation, speed
&$<$1 $^{\circ}$, $<$ 1 mm, $<$ 1s \\ \noalign{\smallskip}
%-----------------
Sadeghi \cite{paper56} &2008  &Bezier curves + GRT  & robot-assisted US images on agar phantom designed
&Speed, mean of needle axis and tip detection error
&- \\ \noalign{\smallskip}
%-----------------
Zhou \cite{paper44}  &2008      &3DIHT + coarse-fine search strategy &3D US captured from water phantom 
&Accuracy of orientation deviation, position deviation, speed
& 1.58 $^{\circ}$, 1.92 mm, 1.76s \\ \noalign{\smallskip}
%-----------------
Aboofazeli \cite{paper136} &2009  &Projection of 3D on 2D image &3D US images of phantom from 3D motorized curvilinear probe 
&Average accuracy of needle tip location
&$<$ 2.8mm \\ \noalign{\smallskip}
%-----------------
Hartmann \cite{paper114} &2009  &3DHT + thresholding   &14 in-vivo 3D US images
&Accuracy of angles between manual and method,
&mean of 2.1 $^{\circ}$ \\ \noalign{\smallskip}
%-----------------
Zhao \cite{paper30}       &2009     &3D Phase-Grouping                 &3D US data from rotational scanning approach on agar and water phantoms
&Angular deviation, Position deviation, speed
&- \\ \noalign{\smallskip}
%-----------------
Adebar \cite{paper102}  &2013    &Doppler Ultrasound and Vibration   &3D US data of liver ex-vivo and PVC phantoms from a Convex mechanical 3D transducer 
&Needle localization and segmentation error
&-\\ \noalign{\smallskip}
%-----------------
Greer \cite{paper93}     &2014     &Doppler Ultrasound and Vibration + B-mode US   &3D US data of liver ex-vivo from tracked linear 2D US Probe
&Accuracy of needle location and its tip
&0.38$\pm$0.27mm, 0.71$\pm$0.55mm \\ \noalign{\smallskip}
%-----------------
\noalign{\smallskip}\hline
\end{tabular}
\end{table*}
% -------- end table 3 --------------------

%============ Kidney ================
\subsection{Kidney}
Segmentation of kidney is a challenging kinds of image processing because kidneys are protected by the lower ribs and in many cases surrounded by fatty tissues. In addition, shadowing, noise and attenuation of US images are very common in this case. Because renal shapes have a bean-shape structure and easy to find with human eyes, to overcome imaging problems, many modalities used reference shapes, then classifiers are applied to detect organ tissues from non-organ tissue. 
\newline In general, non-lesion parts of the Kidney could be useful for survival and better life in body of patients with cancerous situations. Therefore, surgeons need to know about the localization of cancerous tissues as one of the most important information during the operation. Thus, segmentation of kidney from the 3D US images in real-time operations is a vital task and significantly improve the operation results. \newline Ahmad et al. \cite{paper25} proposed two guided and unguided methods for segmentation of Kidney 3D data which are acquired with optical tracked freehand reconstruction method. In unguided modality, for each 2D image, operator have to specify points of tumor boundary, then Discrete Dynamic Contour (DDC) approach detect the kidney tumors. In guided method, 3D segmented tumor is reconstructed by just one 2D slice after adding the kidney cancer shape information which is usually spherical, egg-shaped and have symmetry.
\newline In Contrast Enhanced Ultrasound (CEUS) imaging modality, Gas-filled micro-bubbles which have ability of high reflect sound waves are injected to desired tissue to enhance US image contrast. Prevost et al. \cite{paper63} and \cite{paper89} proposed a method that kidney center, size and orientation was detected and segmented in real-time CEUS by a novel automatic ellipsoid detector algorithm. Then template deformable model technique was used for visualization of the kidney and after finding the contours user can have a modification on it. Noll et al. \cite{paper91} used level-set method and fast marching algorithm to find kidney position and visualization using its shape and intensities as a prior knowledge.  
\newline In \cite{paper103} combination of 3D CEUS and 3D US images are used for decision forest and random forest classifiers to make a map of probabilistic classes of kidney location. These classes are applied to a joint methods that attempt to segment a similar object in several images and register them using ellipsoid detector technique and template deformable model. Kidney lesions like cysts and necrotic volumes are segmented by Prevost et al. \cite{paper33} designed for contrast-enhanced US (CEUS) imaging. After applying a fast anisotropic filter to images, fast radial symmetry transform (TRST) searches for dark spherical shapes and maximum likelihood of pixels is considered as the lesion center. Then fast marching algorithm and front propagation method obtain the whole lesion region. 
\newline For segmentation of cancer from healthy and pathological kidneys 3D data, Cerrolaza et al. \cite{paper70} first utilized a weighted statistical shape model as an enhancement of 3D US images respect to the propagation direction of the sound waves, then a new active shape model (multi scale Gabor-based Appearance texture Model (GAM)) applied to reduce speckle noises and detect the kidney contours at different levels of resolution. The similar research group \cite{paper96} used a positive delta shape detector and active contour-based formulation to determine the position of fluid collecting system and its surrounded fats in the kidney, then GAM is applied to completely delineate renal tissues of the patient. 
\newline Marsousi et al. in \cite{paper139} introduced a new automatic kidney outline detection method using deformable model and level-set propagation and applied it on Morison's pouch ultrasound images. Prior knowledge about various shapes of kidney as a probabilistic kidney shape model (PKSM) is utilized for initialization of the deformable model.
For elimination of manual initialization they also proposed an atlas-based 3D segmentation method \cite{paper77} using texture and shape information of the kidney. An Spatially Aligned Neural Network (SANN) classifier registers 3D US images with their ground-truth data on reference volume in training stage. These registered training volumes are used to generate the atlas database. Then in the the next stage, a feature-based rigid registration is used to fit input images on atlas volumes and SANNs are applied to classify voxels into kidney and non-kidney candidates based on texture information. Finally, the region-based level-set method is used for segmentation of the kidney. A similar approach for automatic boundary initialization using a database through support vector machine (SVM) and model-based deformation is proposed by Ardon et al.\cite{paper73}.   
\begin{figure}
\includegraphics[width=0.48\textwidth]{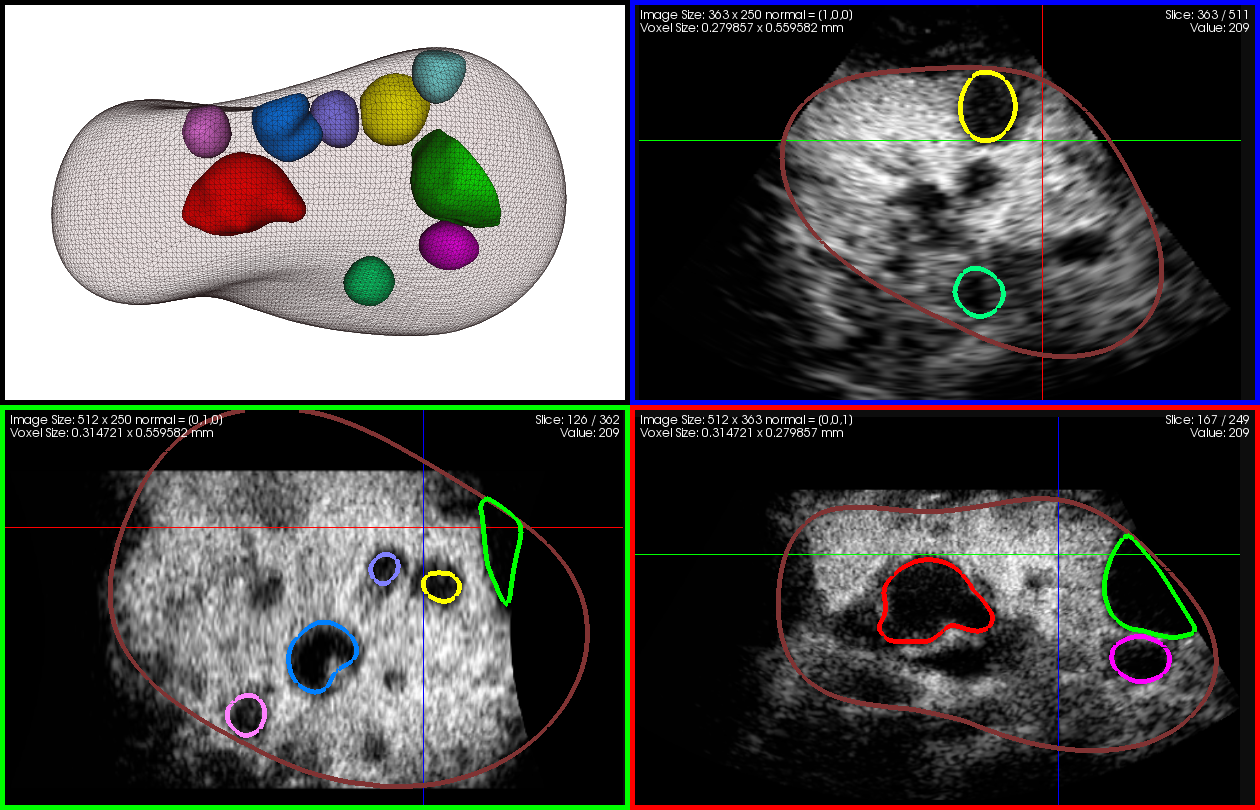}
\caption{3D visualization of kidney lesions \cite{paper33}}
\label{fig:2}       % Give a unique label
\end{figure}
Kidney segmentation methods are illustrated in \ref{table4} and table keywords are: Dice's Coefficient: DSC, PSD: the Point-to-Surface Distance, RVD: Relative Volume Difference, SPSD: Symmetric Point-to-Surface Distance, ACC: Accuracy Measure, MD: Mean Distance. 
%----------  Table 4 -----------------
\begin{table*}[t]
\caption{Needle segmentation methods}
\label{table4}  
\begin{tabular}{m{2cm}m{1cm}m{3cm}m{2.5cm}m{4cm}m{3cm}}
\hline
\noalign{\smallskip}
%-----------------
Reference                      &Year         &Modalities                               &Samples Acquisition                    
&Evaluation Methods                                                                
& Values\\ \hline\noalign{\smallskip}
%-----------------
Ahmad \cite{paper25}   &2006        &Guided and Unguided DDC       &3D data of phantom from Optical tracking + Freehand 2D prob
&Hausdorff distance  
& \\ \noalign{\smallskip}
%-----------------
Prevost \cite{paper63}, \cite{paper89}  &2012, 2014       &Ellipsoid detector algorithm + deformable model   &CEUS 3D volumes,  
&DSC 
&median of 0.84 \\ \noalign{\smallskip}
%-----------------
Prevost \cite{paper33}  &2012     &TRST + Front Propagation and Marching algorithm  &8 3D volume data
&-
&- \\ \noalign{\smallskip}
%-----------------
Prevost \cite{paper103}  &2013   &Ellipsoid detection + Deformable Model + Random Forest Classifier &64 couples of CEUS and US volumes
&DSC
&median 0.81 in CEUS, 0.78 in US \\ \noalign{\smallskip}
%-----------------
Marsousi \cite{paper139}  &2014    &ASM + Level-set    &14 3D US volumes 
&DSC, t-test
&0.6552$\pm$0.0595, 0.000032 \\ \noalign{\smallskip}
%-----------------
Noll \cite{paper91}  &2014	&Marching algorithm + Level-set Method  &56 3D US data from 8 patients 
&-   
&- \\ \noalign{\smallskip}
%-----------------
Cerrolaza \cite{paper70}  &2014	&3D GAM + ASM + weighted static shape model  &14 3D US pediatric right kidneys images 
&DSC, PSD, RVD
&0.85$\pm$0.03, 4.07$\pm$1.11 mm, 0.12$\pm$0.08 \\ \noalign{\smallskip}
%-----------------
Cerrolaza \cite{paper96}   &2015    &3D GAM + ASM + positive delta detector  &13 3D US data 
&DSC, SPSD
&0.75$\pm$0.08, 0.98$\pm$0.27 mm \\ \noalign{\smallskip}
%-----------------
Marsousi \cite{paper77}  &2015    &Affine registration + SANN + level-set  &36 3D US images
&DSC, ACC, MD
&0.51$\pm$0.17, 94.01$\pm$1.93, 3.84$\pm$2.12 \\ \noalign{\smallskip}
%-----------------
Ardon \cite{paper73}  &2015     &Deformable model + SVM + editing stage   &480 3D images (360 for testing, 120 for learning)
&DSC
&median 0.91 \\ \noalign{\smallskip}
%-----------------
\noalign{\smallskip}\hline
\end{tabular}
\end{table*}
% -------- end table 4 --------------------

%============ Embryo and Fetus ================
\subsection{Embryo and Fetus}

Today, by development of medical imaging, it is possible to monitor fetal growth and assess its development during pregnancy. Use of three dimensional ultrasound (3D US) is preferred over other tomographic modalities due to its real-time representation, non-invasiveness and low cost. Therefore, 2D and 3D medical ultrasound imaging are the main techniques being used during pregnancy. It is routinely used for well being assessment of the fetus by measuring some characteristics such as cerebellum volume, fetal biometric, the bi-parietal diameter, morphology analysis or the crown-rump length. Although 3D US imaging is a perfect method for fetus analysis but the image quality and the resolution of the images is often low with noisy patterns so that the automatic segmentation of fetal structures become a challenging problem.
\newline Anquez et al. \cite{paper54} proposed a method using intensities of the pixels associate with fetal, maternal tissues and amniotic fluid which are represented with Rayleigh and exponential distribution and applied these data as a prior information to level-set deformable model to segment regions of interest. The similar research group \cite{paper141} introduced a complementary version of their previous method by adding 2 more distributions. They considered segmentation as a classification which attempts to find the optimal classes between the amniotic fluid area and the embryonic tissues. 
\newline Based on aforementioned method, Dahdouh et al. \cite{paper67} and \cite{paper11} applied a shape prior constraint which is coded with Legendre moments as a level-set segmentation approach and tissue-specific parametric intensity distribution modeling to segment 3D fetal US images. In both methods a database of fetus back shapes also applied to outputs for better segmentation of the fetus from connected structures such as the uterus wall.
\newline In \cite{paper60} a 3D Point Distribution Model (PDM) has been used for representing the cerebellum which is automatically adjusted to a 3D US volume using a genetic algorithm (in \cite{paper80} they used Nelder-Mean simplex algorithm) for optimizing model fitting objective function. The same authors group in \cite{paper15} have introduced their automatic segmentation by using another objective functions which is intended to find minimum differences between PDM and the surface of the cerebellum in an US volume. They also used prior knowledge of the anatomy of the cerebellum which manually is obtained by expert obstetricians in training stages. 
\newline A novel method for segmentation of the fetal brain whole structure in 3D US images was built by Yaqub et al. \cite{paper90} and \cite{paper90-ref4} as a classification problem. They used Random Decision Forests (RDF) and fast-weighted RDF classifier which is a new machine learning technique for guiding segmentation process. In \cite{paper90-ref8} a similar method have been used for delineation of the myocardium. In \cite{paper69} the authors first registered the high frequency US datasets with prior manual reference segmented training data from mouse embryo brain ventricle, then an active shape model ASM with growing and shrinking region method have been used to get the final result. The same authors group in \cite{paper78-ref4} introduced a new automatic method (nested graph cut (NGC)) which can segment multiple nested objects such as brain ventricles (BVs), head, amniotic fluid and uterus precisely. They also proposed in \cite{paper78} a modality by extracting the skeleton of BV and decomposing it into five regions (fourth ventricle, aqueduct, third ventricle and two lateral ventricles) in order to characterized BVs shape variation. 
\newline Liu et al. \cite{paper117} for segmentation and volume measurement of fetal cerebellum on 3D US added an external energy term to the active surface model (ASM) using directional phase symmetry of the image which increased the method performance in terms of noise cancellation and continuity of image features edges. Bello-Munoz et al. \cite{paper83} calculate the volume of fetal cerebral structures from 3D ultrasound used segmentation template-based system. Qiu et al. \cite{paper71} used a semi-automatic 3D US data segmentation of cerebral ventricles in preterm neonates for diagnosis of intra-ventricular hemorrhage which an convex optimization algorithm attempts to find the minimum length boundary and energy function of the region.

%----------  Table 5 -----------------
\begin{table*}
\caption{Embryo and Fetus segmentation methods}
\label{table5}  
\begin{tabular}{m{2cm}m{1cm}m{3cm}m{2.5cm}m{4cm}m{3cm}}
\hline
\noalign{\smallskip}
%-----------------
Reference                      &Year         &Modalities                               &Samples Acquisition                    
&Evaluation Methods                                                                
& Values\\ \hline\noalign{\smallskip}
%-----------------
Anquez \cite{paper54}   &2008        &Optimum Deformable model + Level-set + 2 Pdf Distribution              &One 3D dataset
&Classification error rate
&- \\ \noalign{\smallskip}
%-----------------
Anquez \cite{paper141}   &Optimum Deformable model + Level-set + 4 Pdf Distributions    &One 3D dataset
&Sensitivity analysis with respect to initialization and parameters, Overlap measurement 
&- \\ \noalign{\smallskip}
%-----------------

\noalign{\smallskip}\hline
\end{tabular}
\end{table*}
% -------- end table 5 --------------------

%============ Cardiovascular and Carotid arteries ================
\subsection{Cardiovascular and Carotid arteries}

Human head blood is supplied by a major vessels named Carotid from the heart and usually are subject to build-up of plaque which may break loose and follow the blood flow into the brain and cause stroke. Human heart consists two main chambers with responsibility for pumping the blood. Left ventricle is more stronger than the right one and pump out the blood while the right ventricle pump blood into the heart. Cardiovascular diseases have become increasingly common and working process of these ventricles are vital for whole body performance. 
\newline Stroke often is a side-effect of Atherosclerosis and lead to a huge amount of mortality in the developed countries \cite{paper106}. One of the most common cause for occurrence of stroke is the severity of atherosclerosis at the carotid artery bifurcation. Poor function of heart sections is also a heart attack cause in elders. To minimized the diseases which are linked to carotid obstruction, hearth performance and better diagnosis of them physicians need to precisely measure the volume and visualize the shape of them. 
\newline Despite the US imaging weaknesses such as noise, low contrast, acoustic shadowing, a small field of view in 2D US, it plays an important role in the analysis of afore-mention regions function and assessment since it allows a real-time observation. Thus accurate segmentation of these regions from 3D US images could substantially support clinical diagnosis and prevention of heart disease and stroke. 
Gill et al. \cite{paper46} and \cite{paper18} proposed a semi-automatic method for segmentation of carotid artery lumen from freehand 3D US images by using a deformable model. they used a balloon model (like \cite{paper109}) for initialization and compared their method with manually segmentation results. In \cite{paper119} the same research group applied an image-based force to further deform the dynamic balloon model and obtain a better localization of the vessel wall boundary. 
\newline Li et al. \cite{paper22} introduced a new method with geometrically deformable model (GDM) which is a developed version of the active contour model with less initial parameters. In their automatic segmentation method the user need to set up only two initial contours and four parameters. The similar method with some seed user initialization have designed by \cite{paper22-ref8}. Lumen delineation have been performed in \cite{paper74} three manually-placed seed points and using a graph search algorithm following a level set method which is applied on contrast enhanced 3D US images. 
\newline Hossain et al. \cite{paper36}, \cite{paper40}, \cite{paper111}, \cite{paper3} add a novel stopping criterion and initialization stage to a distance regularized level-set with internal and external energy for lumen semi-automatic segmentation in 3D US. User must define some specific slides contours as initial step and difference between Hausdorff distance (MHD) between contours at successive iterations and a stopping boundary is used as a stopping criterion. In \cite{paper105}, a graph based method (Surface Graph Cuts) is used for delineating carotid arteries lumen (especially estimation of bifurcation point position) from free-hand 3D US slices. Lumen centerline defined by user and is extracted and tracked through a ellipse fitting in transversal 2D cross-sections slice. Graph Cut method along with a limited user initialization also have exploited for segmentation of heart annulus in \cite{paper23}. Chalopin et al. \cite{paper84} demonstrated the performance of a semi-automatic approach for intra-operative segmentation of cerebral vascular network from contrast enhanced 3D US data using a model-based multi-scale analysis to estimate center position of cylinder model of tubular structures. 
\newline To segment the heart left ventricle (LV) and atrium from 3D US volumes, Juang et al. \cite{paper61} have proposed an automatic segmentation graph-based method in cylindrical coordinate space. The radial symmetry of the above-mentioned regions is utilized to find a central axis for cylindrical 3D volumes. For similar purpose, Santiago et al. \cite{paper88} presented a new deformable model based on Probabilistic Data Association Filter for removing artifacts from surrounded organs. For having a better image boundary features to use in LV segmentation, authors in \cite{paper94} have used Probabilistic Edge Map (PEM) as a image boundary representation technique which is powerful enough to extract edge borders from 3D US slices. They also have taken advantage of a multi-atlas LV segmentation as a reference. 
\newline Keraudren et al. \cite{paper112} proposed a new automatic method for LV delineation using Random Forest classifiers (RF) which is a machine learning modality that averages the results of decision trees trained on random subsets of the training dataset. Segmentation of the right ventricle of the heart using model-based approach in real-time is studied in details by \cite{paper108}. Nillesen et al. \cite{paper57} have attempted to present a method which segment anatomical structures in children with congenital heart disease without using prior knowledge on the shape of the heart. They tack advantage of radio frequency (RF) signals correlation respect to myocardial and endocardial tissue than moving blood regions in heart and used it as a feature for distinguishing between blood and the heart muscle surfaces. The same research group in \cite{paper100} also utilized maximum temporal cross-correlation values as an additional external force in a deformable model approach. Matinfar and Zagrochev \cite{paper72} proposed a triangular shape deformable model segmentation of pediatric aortic root 3D US images (with internal and external energy forces).  A combination of common 3D US reconstruction methods from 2D slices and segmentation of that images are presented in \cite{paper106}. 

Dice Similarity Coefficient (DSC), Mean absolute distances (MAD), Maximum absolute distances (MAXD), Standard Deviation (SD), Modified Hausdorff Distance (MHD), Hausdorff Distance (HD)

\begin{table*}
\caption{Embryo and Fetus segmentation methods}
\label{table5}  
\begin{tabular}{m{1.25cm}cm{2.5cm}cm{3cm}m{2.5cm}m{3cm}}
\hline
\noalign{\smallskip}
%-----------------
Reference                      &Year         &Modalities                     &A/M  &Samples Acquisition                    
&Evaluation Methods                                                                
& Values\\ \hline\noalign{\smallskip}
%-----------------
Gill et al \cite{paper18}, \cite{paper46}, \cite{paper119}   &1999, 2000             &deformable model     &M         &freehand 3D US image of a human carotid bifurcation
&-
&- \\ \hline\noalign{\smallskip}
%-----------------
Zahalka et al \cite{paper22-ref8}   &2001             &geometrically
deformable model (GDM)       &A        &3D US volume of a stenosed
vessel phantoms + two mechanical scanned 3D vivo samples
&-
&- \\ \hline\noalign{\smallskip}
%-----------------
Li et al. \cite{paper22}                    &2002         &Geometrically Deformable
Model (GDM) with automatic merge function                     &A   &simulated 3D images
&Volume Error                   
& Less than 1$\%$\\ \hline\noalign{\smallskip}
%-----------------
Krissian et al. \cite{paper51}          &2003         &model-based multi-scale detection                     &M   &3D US of a procine aorta
&-              
&-\\ \hline\noalign{\smallskip}
%-----------------
Hold et al. \cite{paper110}          &2007         &region growing     &M   &3D freehand ultrasound dataset of the knee
&SD of the estimated center               
&smaller than $5\%$\\ \hline\noalign{\smallskip}
%-----------------
Nillesen et al. \cite{paper100}, \cite{paper57}          &2009         &gradient-based deformable + maximum cross-correlation values and adaptive mean squares (AMS) filter values as external forces                  &A   & full volume images (Philips, iE33)
of four healthy children
&mismatch ratio (Dice)             
&-\\ \hline\noalign{\smallskip}
%-----------------
Schneider et al. \cite{paper23}          &2010         &Graph cuts and deformable surface                   &A   &3D images of the mitral valve
&SD of mean normalized distances and RMS difference in millimetres 
&1.11$\pm$0.19, 1.81$\pm$0.78\\ \hline\noalign{\smallskip}
%-----------------
Ukwatta et al. \cite{paper32}          &2011         &Level-set method         &M   
&3D US images of 30 patients with carotid stenosis of $60\%$ or more          
&DSC, MAD, MAXD, Volume differences     
&$95.2\%\pm1.6\%$ and $94.3\%\pm2.6\%$, $0.3\pm0.1mm$ and $0.2\pm0.1mm$, $0.8\pm0.4mm$ and $0.6\pm0.3mm$, $4.2\%\pm3.1\%$ and $3.4\%\pm2.6\%$ \\ \hline\noalign{\smallskip}
%-----------------
Juang et al. \cite{paper61}          &2011         &Graph cuts and the radial symmetry transform                    &A   &3D US intra-operative patient data
&difference error              
&3.41$\pm$4.58 pixels or 2.39$\pm$3.21 mm\\ \hline\noalign{\smallskip}
%-----------------
Yang et al. \cite{paper34}          &2012         &Active shape models (ASMs)  &M   &68 3D US volume data acquired from the left and right carotid arteries of seventeen patients
&DSC, MAD, MAXD             
&$93.6\%\pm2.6\%$ and $91.8\%\pm3.5\%$, $0.28\pm0.17mm$ and $0.34\pm0.19mm$, $0.87\pm0.37mm$ and $0.74\pm0.49mm$,\\ \hline\noalign{\smallskip}
%-----------------
Wang et al. \cite{paper64}          &2012         &Region growing + Marching cubes + deformable model                    &M   &40 slices of 2D image from 3D US image
&-             
&-\\ \hline\noalign{\smallskip}
%-----------------
Arias Lorza et al. \cite{paper105}          &2013         &Surface graph-cut                  &M   &3D freehand US images on the neck
&DSC             
&$84\%$ for healthy volunteers and $66.7\%$ patient data\\ \hline\noalign{\smallskip}
%-----------------
Murad Hossain et al. \cite{paper36}          &2013         &Distance regularized level set + a novel initialization and stopping criteria       &M   &3D US image
&DSC, MHD, HD              
&-\\ \hline\noalign{\smallskip}
%-----------------
Santiago et al. \cite{paper88}          &2013         &deformable model                     &M   &3D US of Left Ventricle
&Average distance, Average error              
&-\\ \hline\noalign{\smallskip}
%-----------------
\end{tabular}
\end{table*}
\begin{table*}
\begin{tabular}{m{1.25cm}cm{2.5cm}cm{3cm}m{2.5cm}m{3cm}}

Chalopin et al. \cite{paper84}          &2012         &Geometric deformation model       &M   &intra-operative contrast-enhanced 3D US angiographic images of 3 patients brain + phantom
&Mean radius, Mean cross-section area,              
&-\\ \hline\noalign{\smallskip}
%-----------------
Keraudren et al. \cite{paper112}          &2014         &Autocontext Random Forests          &A   &Images are used in MICCAI 2014 Challenge
&mean DSC            
&$86.4\%$\\ \hline\noalign{\smallskip}
%-----------------
Murad Hossain et al. \cite{paper40} \cite{paper3}          &2014         &Level set method + edge and region based energy           &M     &3D US of 5 subjects with carotid stenosis more than $50\%$
&DSC, HD, MHD              
&-\\ \hline\noalign{\smallskip}
%-----------------
Matinfar et al. \cite{paper72}          &2014         &model-based + non-rigid registration                 &M   &Image sequences of pediatric 3D US data
&-            
&-\\ \hline\noalign{\smallskip}
%-----------------
Oktay et al. \cite{paper94}          &2015         &Multi-atlas + probabilistic edge map (PEM) registration                   &A   &3D US images of MICCAI 2014 CETUS challenge
&Mean distance, HD, DSC              
&-\\ \hline\noalign{\smallskip}
%-----------------
Cao et al. \cite{paper74}          &2015         &Graph search + level-set + gradient concentration calculations      &M   & 35 3D contrast enhanced US images acquired from 7 patients 
&correlation score and coefficient               
&-\\ \hline\noalign{\smallskip}
%-----------------

\noalign{\smallskip}\hline
\end{tabular}
\end{table*}
% -------- end table 5 --------------------

%============ Miscellaneous clinical purposes ================
\subsection{Miscellaneous clinical purposes}
In the past section, we reviewed segmentation approaches in common medical application classes using 3D US images. Studies are not limited to the above-mentioned medical applications and recently usage of 3D US images could be seen everywhere in science. For concluding our taxonomy in this section some other medical 3D US segmentation papers will be discussed. 
\newline Chalana et al. \cite{paper19} introduce a method for delineation of fluid-filled structures such as the bladder using optimal path-finding algorithm to enable the measurement of volumes of these organs. Ovarian follicle shapes assessment is important for human fertilization and a dominant follicle could be found to have ovulation power by observation of its growth. Cigale and Zazula \cite{paper97} proposed a new method for ovarian segmenting from 3D US data using continuous wavelet transform (CWT). Authors in \cite{paper58} utilized a probabilistic framework based combination of entire ovary as a global feature and each follicle context as a local feature information to detect follicle candidates. Gasnier et al. \cite{paper59} presented a new segmentation approach by improving the quality of 3D US images of tumor tissue vessels utilizing contrast enhanced technique. 
\newline In \cite{paper85} authors have proposed a level set method based on a ridge detector for 3D US segmentation of the Distal Femur. 
Because of the major effect of chest wall shadowing on breast US images, Huisman et al. \cite{paper98} presented a breast tissue segmentation method from 3D US data using combination of a deformable volume model which is optimized by simplex optimization method and Hessian rib shadow enhancement filter. Tan et al. \cite{paper10} automatic method segmentation used approximate cylinder model of chest wall and intensity features classification to detect cancer tissues candidates. Hacihaliloglu et al. \cite{paper99} presented a Bone Segmentation and Fracture Detection method with utilizing signal local phase symmetry features and 3D Log-Gabor filters which could discriminate bony from non-bony structures in 3D US data. In \cite{paper38}, Hessian matrix applied to a semi-automatic segmentation of large bones in freehand 3D US images to enhance the bone surface.
\newline Supervised segmentation of infant hip dysplasia based on optimized graph search have been proposed by \cite{paper81} which slice contours are specified by user as a curve passing through points in the graph and for making surface model are interpolated over the 3D volume.
Combination of 3D US and Computer Tomography (CT) for delineation of eye ball and lens structure in radiotherapy planning of retinoblastoma have been utilized in \cite{paper107} using 3D active contour based approach and geometric deformable model with prior knowledge of the eye anatomy. 3D Endobronchial ultrasound (EBUS) data which have been reconstructed from 2D slices used for segmentation of lung-cancer tissue using a graph-search algorithm \cite{paper48}. Lee et al. \cite{paper39} demonstrated the performance of a deformable model using a hybrid edge and region information for liver cancer tumor segmentation from 3D US images. In their method four features from segmented tumor is extracted and support vector machine classifies these features as an supervised optimization procedure to tumor and non-tumor regions. 
\newline Recently, researchers have discovered that Parkinson's Disease could be a consequence of degeneration of some nerve cells in mid-brain and transcranial ultrasound (TC US) imaging is a common way for visualization of this region in human mind. Ahmadi et al. \cite{paper101} developed a new 3D Mid-brain segmentation method from TC US data using a statistical active polyhedron shape model. The same research group \cite{paper95} trained a new implementation of Random Forests classifier (Hough Forests) by a set of 3D US volume data and test it as a automatic segmentation method for discrimination of mid-brain, prostate and heart tissues. In \cite{paper104} mid-brain structure have been segmented form 3D US images by using fusion of experts manual segmented results in training stage of a random forest classifier. 
Delineation method for skin cancer from High frequency 3D US images was introduced in \cite{paper13} which a level-set algorithm used for discrimination between tumors and tissues. The authors revealed that Parzen non-parametric method could be estimate the log-likelihood of contours according to the regions distribution. Linguraru et al. \cite{paper26} designed an semi-automatic texture-based 3D US segmentation algorithm for discrimination between surgery instruments, blood and tissue in real-time intracardiac procedures. They applied expectation maximization algorithm to calculate statistical distribution of each object classes to distinguish each region voxels from neighborhoods, then applied watershed transform to corrects the segmentation errors. Olivier et al. \cite{paper24} presented texture-based methods for skin segmentation from 3D US data using a multi resolution scheme for volumetric texture and an supervised binary classifier with manual initialization. 

%============ Conclusion ================
\section{Conclusion}
In this paper, we gathered many methods of 3D Ultrasound image segmentation focusing on clinical applications. Although many approaches have been done for 3D image segmentation but this field is still active and there is not an standard 3D segmentation method in image processing for any purposes. Thus, for a new problem, it is important to consider available methods and choose the best suitable one which is capable to solve that problem well. For this reason we list all the 3D methods in tables and one can compare then in terms of interactivity, and powerfulness using different evaluation methods. 
\newline Surveying the methods for 3D image segmentation shows that the classical 2D techniques can also be powerful and used for 3D cases. In the other hand, in the most cases, the 3D US image segmentation methods are the extended version of corresponding 2D methods. Thus, new algorithms that works from the scratch in 3D or also 4D domain can be considered as future works. 
\newline Like 2D US segmentation approaches that are dependent on image resolution and contrast, 3D segmentation techniques are linked to the quality of the 3D reconstructed image from 2D slices and voxel resolution. Accordingly, 3D US image segmentation must be always considered with image reconstruction algorithm. Here we have just presented a review on general properties of the techniques. We show their applications with focus on clinical cases and details of each methods are left to the reader. 

%=====================================
%\begin{acknowledgements}
%\end{acknowledgements}
\bibliographystyle{spphys}       % APS-like style for physics
\bibliography{template}   % name your BibTeX data base

\begin{thebibliography}{100}
\providecommand{\url}[1]{{#1}}
\providecommand{\urlprefix}{URL }
\expandafter\ifx\csname urlstyle\endcsname\relax
  \providecommand{\doi}[1]{DOI \discretionary{}{}{}#1}\else
  \providecommand{\doi}{DOI \discretionary{}{}{}\begingroup
  \urlstyle{rm}\Url}\fi

\bibitem{ProstateStatistic}
P.D. Baade, D.R. Youlden, L.J. Krnjacki, Molecular nutrition \& food research
  \textbf{53}(2), 171 (2009)

\bibitem{paper1}
N.~Hu, D.B. Downey, A.~Fenster, H.M. Ladak, Medical physics \textbf{30}(7),
  1648 (2003)

\bibitem{paper130}
I.B. Tutar, S.D. Pathak, L.~Gong, P.S. Cho, K.~Wallner, Y.~Kim, Medical
  Imaging, IEEE Transactions on \textbf{25}(12), 1645 (2006)

\bibitem{paper122}
A.~Ghanei, H.~Soltanian-Zadeh, A.~Ratkewicz, F.F. Yin, Medical Physics
  \textbf{28}(10), 2147 (2001)

\bibitem{paper49}
N.~Hu, D.B. Downey, A.~Fenster, H.M. Ladak, in \emph{Biomedical Imaging, 2002.
  Proceedings. 2002 IEEE International Symposium on} (IEEE, 2002), pp. 613--616

\bibitem{paper120}
Y.~Wang, H.N. Cardinal, D.B. Downey, A.~Fenster, Medical physics
  \textbf{30}(5), 887 (2003)

\bibitem{paper124}
M.~Ding, C.~Chen, Y.~Wang, I.~Gyacskov, A.~Fenster, in \emph{Medical Imaging
  2003} (International Society for Optics and Photonics, 2003), pp. 69--76

\bibitem{paper123}
M.~Ding, I.~Gyacskov, X.~Yuan, M.~Drangova, A.~Fenster, in \emph{Medical
  Imaging 2004} (International Society for Optics and Photonics, 2004), pp.
  151--160

\bibitem{paper129}
M.~Ding, B.~Chiu, I.~Gyacskov, X.~Yuan, M.~Drangova, D.B. Downey, A.~Fenster,
  Medical physics \textbf{34}(11), 4109 (2007)

\bibitem{paper21}
H.M. Ladak, M.~Ding, Y.~Wang, N.~Hu, D.B. Downey, A.~Fenster, in \emph{Medical
  Imaging 2004} (International Society for Optics and Photonics, 2004), pp.
  1403--1410

\bibitem{paper125}
S.~Fan, L.K. Voon, N.W. Sing, in \emph{Medical Image Computing and
  Computer-Assisted Intervention—MICCAI 2002} (Springer, 2002), pp. 389--396

\bibitem{paper127}
F.~Wang, J.~Suri, A.~Fenster, in \emph{Engineering in Medicine and Biology
  Society, 2006. EMBS'06. 28th Annual International Conference of the IEEE}
  (IEEE, 2006), pp. 2341--2344

\bibitem{paper121}
W.~Qiu, J.~Yuan, E.~Ukwatta, D.~Tessier, A.~Fenster, Medical physics
  \textbf{40}(7), 072903 (2013)

\bibitem{paper131}
W.~Qiu, J.~Yuan, E.~Ukwatta, A.~Fenster, Medical physics \textbf{42}(2), 877
  (2015)

\bibitem{paper53}
A.C. Hodge, H.M. Ladak, in \emph{Engineering in Medicine and Biology Society,
  2006. EMBS'06. 28th Annual International Conference of the IEEE} (IEEE,
  2006), pp. 2337--2340

\bibitem{paper6}
A.C. Hodge, A.~Fenster, D.B. Downey, H.M. Ladak, Computer methods and programs
  in biomedicine \textbf{84}(2), 99 (2006)

\bibitem{paper126-ref14}
D.~Shen, Y.~Zhan, C.~Davatzikos, Medical Imaging, IEEE Transactions on
  \textbf{22}(4), 539 (2003)

\bibitem{paper126}
Y.~Zhan, D.~Shen, Medical Imaging, IEEE Transactions on \textbf{25}(3), 256
  (2006)

\bibitem{paper29}
T.~Heimann, M.~Baumhauer, T.~Simpfend{\"o}rfer, H.P. Meinzer, I.~Wolf, in
  \emph{Medical Imaging} (International Society for Optics and Photonics,
  2008), pp. 69,141P--69,141P

\bibitem{thesis116}
W.~Shao, Prostate segmentation and multimodal registration in 3d ultrasound
  images.
\newblock Ph.D. thesis (2009)

\bibitem{paper45}
X.~Yang, D.~Schuster, V.~Master, P.~Nieh, A.~Fenster, B.~Fei, in \emph{SPIE
  Medical Imaging} (International Society for Optics and Photonics, 2011), pp.
  796,432--796,432

\bibitem{paper35}
X.~Yang, B.~Fei, in \emph{SPIE Medical Imaging} (International Society for
  Optics and Photonics, 2012), pp. 83,162O--83,162O

\bibitem{paper31}
H.~Akbari, X.~Yang, L.V. Halig, B.~Fei, in \emph{Proceedings of SPIE}, vol.
  7962 (NIH Public Access, 2011), vol. 7962, p. 79622K

\bibitem{paper2}
H.~Akbari, B.~Fei, Medical Physics \textbf{39}(6), 2972 (2012)

\bibitem{paper128}
S.S. Mahdavi, N.~Chng, I.~Spadinger, W.J. Morris, S.E. Salcudean, Medical Image
  Analysis \textbf{15}(2), 226 (2011)

\bibitem{paper37}
S.~Nouranian, S.S. Mahdavi, I.~Spadinger, W.J. Morris, S.~Salcudean,
  P.~Abolmaesumi, in \emph{SPIE Medical Imaging} (International Society for
  Optics and Photonics, 2013), pp. 86,710O--86,710O

\bibitem{paper86}
A.~Fenster, M.~Ding, N.~Hu, H.M. Ladak, G.~Li, N.~Cardinal, D.B. Downey, in
  \emph{Computer Vision Beyond the Visible Spectrum} (Springer, 2005), pp.
  241--269

\bibitem{BreastCancerStatistics}
A.~Jemal, F.~Bray, M.M. Center, J.~Ferlay, E.~Ward, D.~Forman, CA: a cancer
  journal for clinicians \textbf{61}(2), 69 (2011)

\bibitem{paper7}
D.R. Chen, R.F. Chang, W.J. Wu, W.K. Moon, W.L. Wu, Ultrasound in medicine \&
  biology \textbf{29}(7), 1017 (2003)

\bibitem{paper17}
J.I. Kwak, M.N. Jung, S.H. Kim, N.C. Kim, in \emph{Medical Imaging 2003}
  (International Society for Optics and Photonics, 2003), pp. 193--200

\bibitem{paper133}
R.F. Chang, W.J. Wu, C.C. Tseng, D.R. Chen, W.K. Moon, Information Technology
  in Biomedicine, IEEE Transactions on \textbf{7}(3), 197 (2003)

\bibitem{paper134}
R.F. Chang, W.J. Wu, W.K. Moon, W.M. Chen, W.~Lee, D.R. Chen, Ultrasound in
  medicine \& biology \textbf{29}(11), 1571 (2003)

\bibitem{paper47}
H.C. Kuo, M.L. Giger, I.~Reiser, K.~Drukker, A.~Edwards, C.A. Sennett, in
  \emph{SPIE Medical Imaging} (International Society for Optics and Photonics,
  2013), pp. 867,025--867,025

\bibitem{paper27}
Q.~Liu, Y.~Ge, Y.~Ou, B.~Cao, in \emph{International Symposium on Multispectral
  Image Processing and Pattern Recognition} (International Society for Optics
  and Photonics, 2007), pp. 67,890D--67,890D

\bibitem{paper55}
S.F. Huang, Y.C. Chen, W.K. Moon, in \emph{Biomedical Imaging: From Nano to
  Macro, 2008. ISBI 2008. 5th IEEE International Symposium on} (IEEE, 2008),
  pp. 1303--1306

\bibitem{paper5}
P.~Gu, W.M. Lee, M.A. Roubidoux, J.~Yuan, X.~Wang, P.L. Carson, Ultrasonics
  \textbf{65}, 51 (2016)

\bibitem{paper43}
T.~Hopp, M.~Zapf, N.~Ruiter, in \emph{SPIE Medical Imaging} (International
  Society for Optics and Photonics, 2014), pp. 90,401R--90,401R

\bibitem{paper113}
H.~Yang, L.~Christopher, N.~Duric, E.~West, P.~Bakic, et~al., in
  \emph{Electro/Information Technology (EIT), 2012 IEEE International
  Conference on} (IEEE, 2012), pp. 1--4

\bibitem{paper132}
L.A. Christopher, E.J. Delp, C.R. Meyer, P.L. Carson, et~al., in \emph{ISBI}
  (2002), pp. 86--89

\bibitem{paper62}
H.~Yang, L.~Christopher, N.~Duric, E.~West, P.~Bakic, et~al., in
  \emph{Electro/Information Technology (EIT), 2012 IEEE International
  Conference on} (IEEE, 2012), pp. 1--4

\bibitem{paper62-2}
H.~Yang,   (2013)

\bibitem{paper16}
M.~Ding, H.N. Cardinal, W.~Guan, A.~Fenster, in \emph{Medical Imaging 2002}
  (International Society for Optics and Photonics, 2002), pp. 65--76

\bibitem{paper79}
M.~Ding, A.~Fenster, Computer Aided Surgery \textbf{9}(5), 193 (2004)

\bibitem{paper118}
M.~Ding, H.N. Cardinal, A.~Fenster, Medical Physics \textbf{30}(2), 222 (2003)

\bibitem{paper4}
M.~Ding, Z.~Wei, L.~Gardi, D.B. Downey, A.~Fenster, Ultrasonics \textbf{44},
  e331 (2006)

\bibitem{paper28}
H.~Zhou, W.~Qiu, M.~Ding, S.~Zhang, in \emph{International Symposium on
  Multispectral Image Processing and Pattern Recognition} (International
  Society for Optics and Photonics, 2007), pp. 67,890R--67,890R

\bibitem{paper44}
H.~Zhou, W.~Qiu, M.~Ding, S.~Zhang, in \emph{Medical Imaging} (International
  Society for Optics and Photonics, 2008), pp. 691,821--691,821

\bibitem{paper114}
P.~Hartmann, M.~Baumhauer, J.~Rassweiler, H.P. Meinzer, in
  \emph{Bildverarbeitung f{\"u}r die Medizin 2009} (Springer, 2009), pp.
  341--345

\bibitem{paper138}
W.~Qiu, M.~Ding, M.~Yuchi, in \emph{Intelligent Networks and Intelligent
  Systems, 2008. ICINIS'08. First International Conference on} (IEEE, 2008),
  pp. 449--452

\bibitem{paper56}
H.R.S. Neshat, R.V. Patel, in \emph{Biomedical Robotics and Biomechatronics,
  2008. BioRob 2008. 2nd IEEE RAS \& EMBS International Conference on} (IEEE,
  2008), pp. 670--675

\bibitem{paper136}
M.~Aboofazeli, P.~Abolmaesumi, P.~Mousavi, G.~Fichtinger, in \emph{Biomedical
  Imaging: From Nano to Macro, 2009. ISBI'09. IEEE International Symposium on}
  (IEEE, 2009), pp. 1067--1070

\bibitem{paper137}
Z.~Wei, L.~Gardi, D.B. Downey, A.~Fenster, in \emph{Biomedical Imaging: Nano to
  Macro, 2004. IEEE International Symposium on} (IEEE, 2004), pp. 960--963

\bibitem{paper30}
S.~Zhao, W.~Qiu, Y.~Ming, M.~Ding, in \emph{Sixth International Symposium on
  Multispectral Image Processing and Pattern Recognition} (International
  Society for Optics and Photonics, 2009), pp. 74,971L--74,971L

\bibitem{paper30-ref5}
J.B. Burns, A.R. Hanson, E.M. Riseman, Pattern Analysis and Machine
  Intelligence, IEEE Transactions on (4), 425 (1986)

\bibitem{paper102}
T.K. Adebar, A.M. Okamura, in \emph{Information Processing in Computer-Assisted
  Interventions} (Springer, 2013), pp. 61--70

\bibitem{paper93}
J.D. Greer, T.K. Adebar, G.L. Hwang, A.M. Okamura, in \emph{Medical Image
  Computing and Computer-Assisted Intervention--MICCAI 2014} (Springer, 2014),
  pp. 381--388

\bibitem{paper25}
A.~Ahmad, D.~Cool, B.H. Chew, S.E. Pautler, T.M. Peters, in \emph{Medical
  Imaging} (International Society for Optics and Photonics, 2006), pp.
  61,410S--61,410S

\bibitem{paper63}
R.~Prevost, B.~Mory, J.M. Correas, L.D. Cohen, R.~Ardon, in \emph{Biomedical
  Imaging (ISBI), 2012 9th IEEE International Symposium on} (IEEE, 2012), pp.
  1559--1562

\bibitem{paper89}
R.~Prevost, B.~Mory, R.~Cuingnet, J.M. Correas, L.D. Cohen, R.~Ardon, in
  \emph{Abdomen and Thoracic Imaging} (Springer, 2014), pp. 37--67

\bibitem{paper91}
M.~Noll, X.~Li, S.~Wesarg, in \emph{Clinical Image-Based Procedures.
  Translational Research in Medical Imaging} (Springer, 2014), pp. 83--90

\bibitem{paper103}
R.~Prevost, R.~Cuingnet, B.~Mory, J.M. Correas, L.D. Cohen, R.~Ardon, in
  \emph{Information Processing in Medical Imaging} (Springer, 2013), pp.
  268--279

\bibitem{paper33}
R.~Prevost, L.D. Cohen, J.M. Corr{\'e}as, R.~Ardon, in \emph{SPIE Medical
  Imaging} (International Society for Optics and Photonics, 2012), pp.
  83,141D--83,141D

\bibitem{paper70}
J.J. Cerrolaza, N.~Safdar, C.A. Peters, E.~Myers, J.~Jago, M.G. Linguraru, in
  \emph{Biomedical Imaging (ISBI), 2014 IEEE 11th International Symposium on}
  (IEEE, 2014), pp. 633--636

\bibitem{paper96}
J.J. Cerrolaza, C.~Meyer, J.~Jago, C.~Peters, M.G. Linguraru, in \emph{Medical
  Image Computing and Computer-Assisted Intervention—MICCAI 2015} (Springer,
  2015), pp. 711--718

\bibitem{paper139}
M.~Marsousi, K.N. Plataniotis, S.~Stergiopoulos, in \emph{Engineering in
  Medicine and Biology Society (EMBC), 2014 36th Annual International
  Conference of the IEEE} (IEEE, 2014), pp. 2890--2894

\bibitem{paper77}
M.~Marsousi, K.N. Plataniotis, S.~Stergiopoulos, in \emph{Engineering in
  Medicine and Biology Society (EMBC), 2015 37th Annual International
  Conference of the IEEE} (IEEE, 2015), pp. 2001--2005

\bibitem{paper73}
R.~Ardon, R.~Cuingnet, K.~Bacchuwar, V.~Auvray, in \emph{Biomedical Imaging
  (ISBI), 2015 IEEE 12th International Symposium on} (IEEE, 2015), pp. 268--271

\bibitem{paper54}
J.~Anquez, E.D. Angelini, I.~Bloch, in \emph{Biomedical Imaging: From Nano to
  Macro, 2008. ISBI 2008. 5th IEEE International Symposium on} (IEEE, 2008),
  pp. 17--20

\bibitem{paper141}
J.~Anquez, E.D. Angelini, G.~Grang{\'e}, I.~Bloch, Biomedical Engineering, IEEE
  Transactions on \textbf{60}(5), 1388 (2013)

\bibitem{paper67}
S.~Dahdouh, A.~Serrurier, G.~Grang{\'e}, E.D. Angelini, I.~Bloch, in
  \emph{Biomedical Imaging (ISBI), 2013 IEEE 10th International Symposium on}
  (IEEE, 2013), pp. 1026--1029

\bibitem{paper11}
S.~Dahdouh, E.D. Angelini, G.~Grang{\'e}, I.~Bloch, Medical image analysis
  (2015)

\bibitem{paper60}
B.G. Becker, F.A. Cosio, M.E.G. Huerta, J.A. Benavides-Serralde, in
  \emph{Engineering in Medicine and Biology Society (EMBC), 2010 Annual
  International Conference of the IEEE} (IEEE, 2010), pp. 4731--4734

\bibitem{paper80}
B.~Guti{\'e}rrez-Becker, F.A. Cos{\'\i}o, M.E.G. Huerta, J.A.
  Benavides-Serralde, L.~Camargo-Mar{\'\i}n, V.M. Ba{\~n}uelos, Medical \&
  biological engineering \& computing \textbf{51}(9), 1021 (2013)

\bibitem{paper15}
B.G. Becker, F.A. Cos{\'\i}o, M.G. Huerta, J.~Benavides-Serralde, in
  \emph{Simposio Mexicano en Cirug{\'\i}a Asistida por Computadora y
  Procesamiento de Im{\'a}genes M{\'e}dicas} (2011)

\bibitem{paper90}
M.~Yaqub, R.~Cuingnet, R.~Napolitano, D.~Roundhill, A.~Papageorghiou, R.~Ardon,
  J.A. Noble, in \emph{Machine Learning in Medical Imaging} (Springer, 2013),
  pp. 25--32

\bibitem{paper90-ref4}
M.~Yaqub, K.~Javaid, C.~Cooper, A.~Noble, in \emph{MICCAI Workshop on Machine
  Learning in Medical Imaging, Toronto, Canada} (2011), pp. 1--8

\bibitem{paper90-ref8}
V.~Lempitsky, M.~Verhoek, J.A. Noble, A.~Blake, in \emph{Functional Imaging and
  Modeling of the Heart} (Springer, 2009), pp. 447--456

\bibitem{paper69}
J.W. Kuo, Y.~Wang, O.~Aristiz{\'a}bal, J.~Ketterling, J.~Mamou, et~al., in
  \emph{Ultrasonics Symposium (IUS), 2013 IEEE International} (IEEE, 2013), pp.
  1781--1784

\bibitem{paper78-ref4}
J.w. Kuo, J.~Mamou, O.~Aristizabal, X.~Zhao, J.~Ketterling, Y.~Wang,   (2015)

\bibitem{paper78}
J.w. Kuo, Y.~Wang, O.~Aristizabal, D.H. Turnbull, J.~Ketterling, J.~Mamou, in
  \emph{Ultrasonics Symposium (IUS), 2015 IEEE International} (IEEE, 2015), pp.
  1--4

\bibitem{paper117}
X.~Liu, J.~Yu, Y.~Wang, P.~Chen, British Journal of Health Informatics and
  Monitoring \textbf{1}(2) (2014)

\bibitem{paper83}
J.~Bello-Munoz, J.~Fernandez, J.~Benavides-Serralde, E.~Hernandez-Andrade,
  E.~Gratacos, Ultrasound in Obstetrics \& Gynecology \textbf{30}(4), 498
  (2007)

\bibitem{paper71}
W.~Qiu, J.~Yuan, J.~Kishimoto, S.~de~Ribaupierre, E.~Ukwatta, A.~Fenster, in
  \emph{Biomedical Imaging (ISBI), 2014 IEEE 11th International Symposium on}
  (IEEE, 2014), pp. 1222--1225

\bibitem{paper106}
E.~Ukwatta, A.~Fenster, in \emph{Advanced Computational Approaches to
  Biomedical Engineering} (Springer, 2014), pp. 131--157

\bibitem{paper46}
J.~Gill, H.~Ladak, D.~Steinman, A.~Fenster, in \emph{[Engineering in Medicine
  and Biology, 1999. 21st Annual Conference and the 1999 Annual Fall Meetring
  of the Biomedical Engineering Society] BMES/EMBS Conference, 1999.
  Proceedings of the First Joint}, vol.~2 (IEEE, 1999), vol.~2, pp. 1146--vol

\bibitem{paper18}
J.D. Gill, H.M. Ladak, D.A. Steinman, A.~Fenster, in \emph{Medical Imaging'99}
  (International Society for Optics and Photonics, 1999), pp. 214--221

\bibitem{paper109}
P.~Mattsson, A.~Eriksson,   (2002)

\bibitem{paper119}
J.D. Gill, H.M. Ladak, D.A. Steinman, A.~Fenster, Medical physics
  \textbf{27}(6), 1333 (2000)

\bibitem{paper22}
X.~Li, Z.~Wang, H.~Lu, Z.~Liang, in \emph{Medical Imaging 2002} (International
  Society for Optics and Photonics, 2002), pp. 1458--1463

\bibitem{paper22-ref8}
A.~Zahalka, A.~Fenster, Physics in medicine and biology \textbf{46}(4), 1321
  (2001)

\bibitem{paper74}
K.~Cao, D.~Padfield, A.~Dentinger, K.~Wallace, D.~Mills, in \emph{Biomedical
  Imaging (ISBI), 2015 IEEE 12th International Symposium on} (IEEE, 2015), pp.
  659--662

\bibitem{paper36}
M.M. Hossain, K.~AlMuhanna, L.~Zhao, B.~Lal, S.~Sikdar, in \emph{SPIE Medical
  Imaging} (International Society for Optics and Photonics, 2013), pp.
  86,694A--86,694A

\bibitem{paper40}
M.M. Hossain, K.~AlMuhanna, L.~Zhao, B.K. Lal, S.~Sikdar, in \emph{SPIE Medical
  Imaging} (International Society for Optics and Photonics, 2014), pp.
  90,344B--90,344B

\bibitem{paper111}
M.M. Hossain, Semiautomatic segmentation of atherosclerotic carotid artery wall
  using 3d ultrasound imaging.
\newblock Ph.D. thesis (2014)

\bibitem{paper3}
M.M. Hossain, K.~AlMuhanna, L.~Zhao, B.K. Lal, S.~Sikdar, Medical physics
  \textbf{42}(4), 2029 (2015)

\bibitem{paper105}
A.M.A. Lorza, D.D. Carvalho, J.~Petersen, A.C. van Dijk, A.~van~der Lugt, W.J.
  Niessen, S.~Klein, M.~de~Bruijne, in \emph{Medical Image Computing and
  Computer-Assisted Intervention--MICCAI 2013} (Springer, 2013), pp. 542--549

\bibitem{paper23}
R.J. Schneider, D.P. Perrin, N.V. Vasilyev, G.R. Marx, P.J. del Nido, R.D.
  Howe, Medical Imaging, IEEE Transactions on \textbf{29}(9), 1676 (2010)

\bibitem{paper84}
C.~Chalopin, K.~Krissian, J.~Meixensberger, A.~M{\"u}ns, F.~Arlt, D.~Lindner,
  Biomedizinische Technik/Biomedical Engineering \textbf{58}(3), 293 (2013)

\bibitem{paper61}
R.~Juang, E.R. McVeigh, B.~Hoffmann, D.~Yuh, P.~Burlina, in \emph{Biomedical
  Imaging: From Nano to Macro, 2011 IEEE International Symposium on} (IEEE,
  2011), pp. 606--609

\bibitem{paper88}
C.~Santiago, J.S. Marques, J.C. Nascimento, in \emph{Mathematical methodologies
  in pattern recognition and machine learning} (Springer, 2013), pp. 163--178

\bibitem{paper94}
O.~Oktay, A.~Gomez, K.~Keraudren, A.~Schuh, W.~Bai, W.~Shi, G.~Penney,
  D.~Rueckert, in \emph{Functional Imaging and Modeling of the Heart}
  (Springer, 2015), pp. 223--230

\bibitem{paper112}
K.~Keraudren, O.~Oktay, W.~Shi, J.V. Hajnal, D.~Rueckert,

\bibitem{paper108}
A.B. Eng{\aa}s,   (2008)

\bibitem{paper57}
M.M. Nillesen, R.G. Lopata, I.H. Gerrits, H.J. Huisman, J.M. Thijssen,
  L.~Kapusta, C.L. de~Korte, in \emph{Biomedical Imaging: From Nano to Macro,
  2009. ISBI'09. IEEE International Symposium on} (IEEE, 2009), pp. 522--525

\bibitem{paper100}
M.M. Nillesen, R.G. Lopata, H.J. Huisman, J.M. Thijssen, L.~Kapusta, C.L.
  de~Korte, in \emph{Medical Image Computing and Computer-Assisted
  Intervention--MICCAI 2009} (Springer, 2009), pp. 927--934

\bibitem{paper72}
B.~Matinfar, L.~Zagrochev, in \emph{Computer Vision and Pattern Recognition
  Workshops (CVPRW), 2014 IEEE Conference on} (IEEE, 2014), pp. 323--328

\bibitem{paper51}
K.~Krissian, J.~Ellsmere, K.~Vosburgh, R.~Kikinis, C.E. Westin, in
  \emph{Engineering in Medicine and Biology Society, 2003. Proceedings of the
  25th Annual International Conference of the IEEE}, vol.~1 (IEEE, 2003),
  vol.~1, pp. 638--641

\bibitem{paper110}
S.~Hold, K.~Hensel, S.~Winter, C.~Dekomien, G.~Schmitz,   (2007)

\bibitem{paper32}
E.~Ukwatta, J.~Awad, A.~Ward, J.~Samarabandu, A.~Krasinski, G.~Parraga,
  A.~Fenster, in \emph{SPIE Medical Imaging} (International Society for Optics
  and Photonics, 2011), pp. 79,630G--79,630G

\bibitem{paper34}
X.~Yang, J.~Jin, W.~He, M.~Yuchi, M.~Ding, in \emph{SPIE Medical Imaging}
  (International Society for Optics and Photonics, 2012), pp. 83,152H--83,152H

\bibitem{paper64}
X.~Wang, Y.~Zhang, in \emph{Information and Automation (ICIA), 2012
  International Conference on} (IEEE, 2012), pp. 698--703

\bibitem{paper19}
V.~Chalana, S.~Dudycha, G.~McMorrow, in \emph{Medical Imaging 2003}
  (International Society for Optics and Photonics, 2003), pp. 500--506

\bibitem{paper97}
B.~Cigale, D.~Zazula, in \emph{11th Mediterranean Conference on Medical and
  Biomedical Engineering and Computing 2007} (Springer, 2007), pp. 1017--1020

\bibitem{paper58}
T.~Chen, W.~Zhang, S.~Good, K.S. Zhou, D.~Comaniciu, in \emph{Computer Vision,
  2009 IEEE 12th International Conference on} (IEEE, 2009), pp. 795--802

\bibitem{paper59}
A.~Gasnier, R.~Ardon, C.~Ciofolo-Veit, E.~Leen, J.M. Corr{\'e}as, in
  \emph{Biomedical Imaging: From Nano to Macro, 2010 IEEE International
  Symposium on} (IEEE, 2010), pp. 300--303

\bibitem{paper85}
C.~H{\"a}nisch, J.~Hsu, K.~Radermacher,

\bibitem{paper98}
H.~Huisman, N.~Karssemeijer, in \emph{Information Processing in Medical
  Imaging} (Springer, 2007), pp. 245--256

\bibitem{paper10}
T.~Tan, B.~Platel, R.M. Mann, H.~Huisman, N.~Karssemeijer, Medical image
  analysis \textbf{17}(8), 1273 (2013)

\bibitem{paper99}
I.~Hacihaliloglu, R.~Abugharbieh, A.~Hodgson, R.~Rohling, in \emph{Medical
  Image Computing and Computer-Assisted Intervention--MICCAI 2008} (Springer,
  2008), pp. 287--295

\bibitem{paper38}
Z.~Fanti, F.~Torres, F.A. Cos{\'\i}o, in \emph{IX International Seminar on
  Medical Information Processing and Analysis} (International Society for
  Optics and Photonics, 2013), pp. 89,220F--89,220F

\bibitem{paper81}
A.R. Hareendranathan, M.~Mabee, K.~Punithakumar, M.~Noga, J.L. Jaremko,
  International journal of computer assisted radiology and surgery pp. 1--12
  (2015)

\bibitem{paper107}
M.B. Cuadra, S.~Gorthi, F.I. Karahanoglu, B.~Paquier, A.~Pica, H.P. Do,
  A.~Balmer, F.~Munier, J.P. Thiran, in \emph{Computational Vision and Medical
  Image Processing} (Springer, 2011), pp. 247--263

\bibitem{paper48}
X.~Zang, M.~Breslav, W.E. Higgins, in \emph{SPIE Medical Imaging}
  (International Society for Optics and Photonics, 2013), pp. 867,505--867,505

\bibitem{paper39}
M.~Lee, J.H. Kim, M.H. Park, Y.H. Kim, Y.K. Seong, B.H. Cho, K.G. Woo, in
  \emph{SPIE Medical Imaging} (International Society for Optics and Photonics,
  2014), pp. 90,341N--90,341N

\bibitem{paper101}
S.A. Ahmadi, M.~Baust, A.~Karamalis, A.~Plate, K.~Boetzel, T.~Klein, N.~Navab,
  in \emph{Medical Image Computing and Computer-Assisted Intervention--MICCAI
  2011} (Springer, 2011), pp. 362--369

\bibitem{paper95}
F.~Milletari, S.A. Ahmadi, C.~Kroll, C.~Hennersperger, F.~Tombari, A.~Shah,
  A.~Plate, K.~Boetzel, N.~Navab, in \emph{Medical Image Computing and
  Computer-Assisted Intervention--MICCAI 2015} (Springer, 2015), pp. 111--118

\bibitem{paper104}
P.~Chatelain, O.~Pauly, L.~Peter, S.A. Ahmadi, A.~Plate, K.~B{\"o}tzel,
  N.~Navab, in \emph{Medical Image Computing and Computer-Assisted
  Intervention--MICCAI 2013} (Springer, 2013), pp. 230--237

\bibitem{paper13}
B.~Sciolla, P.~Ceccato, L.~Cowell, T.~Dambry, B.~Guibert, P.~Delachartre,
  Physics Procedia \textbf{70}, 1177 (2015)

\bibitem{paper26}
M.G. Linguraru, R.D. Howe, in \emph{Medical Imaging} (International Society for
  Optics and Photonics, 2006), pp. 61,443J--61,443J

\bibitem{paper24}
J.~Olivier, L.~Paulhac, \emph{3D Ultrasound Image Segmentation: Interactive
  Texture-Based Approaches} (INTECH Open Access Publisher, 2011)

\end{thebibliography}
%======================================
\end{document}